\documentclass[letterpaper, 10 pt, conference]{ieeeconf}  

\IEEEoverridecommandlockouts                              

\usepackage{amsmath,amsfonts}
\usepackage{array}
\usepackage[caption=false,font=normalsize,labelfont=sf,textfont=sf]{subfig}
\usepackage{textcomp}
\usepackage{stfloats}
\usepackage{url}
\usepackage{verbatim}
\usepackage{graphicx}
\usepackage{stfloats}

\usepackage{color}

\usepackage{cite}
\usepackage{multirow}
\usepackage{makecell}
\usepackage[T1]{fontenc}
\usepackage[T1]{fontenc}
\usepackage{authblk}

\title{\LARGE \bf
Multi-to-Single Knowledge Distillation  for Point Cloud   Semantic Segmentation
}

\author{Shoumeng Qiu$^1$ , Feng Jiang$^1$, Haiqiang Zhang$^2$, Xiangyang Xue$^1$, Jian Pu$^{1\ast}$

\thanks{This work was supported by Shanghai Municipal Science and Technology Major Project (No.2018SHZDZX01), ZJ Lab, and Shanghai Center for Brain Science and Brain-Inspired Technology, NSFC Project (62176061) and STCSM Project (No.22511105000).}
\thanks{$^{1}$Fudan University, $^{2}$Mogo Auto.}

}

\begin{document}

\maketitle
\thispagestyle{empty}
\pagestyle{empty}



\begin{abstract}
3D point cloud semantic segmentation is one of the fundamental tasks for environmental understanding. Although significant progress has been made in recent years, the performance of classes with few examples or few points is still far from satisfactory. In this paper, we propose a novel multi-to-single knowledge distillation framework for the 3D point cloud semantic segmentation task to boost the performance of those hard classes. Instead of fusing all the points of multi-scans directly, only the instances that belong to the previously defined hard classes are fused. To effectively and sufficiently distill valuable knowledge from multi-scans, we leverage a multilevel distillation framework, i.e., feature representation distillation, logit distillation, and affinity distillation. We further develop a novel instance-aware affinity distillation algorithm for capturing high-level structural knowledge to enhance the distillation efficacy for hard classes. Finally, we conduct experiments on the SemanticKITTI dataset, and the results on both the validation and test sets demonstrate that our method yields substantial improvements compared with the baseline method.  The code is available at \Url{https://github.com/skyshoumeng/M2SKD}.


\section{INTRODUCTION}
3D point cloud semantic segmentation aims to classify every point of a given scan into a certain class, which is crucial for various applications such as the navigation of autonomous vehicles \cite{zhang2019review,lu2020deep}. Although significant progress has been made in recent years \cite{hu2020randla,tang2020searching,thomas2019kpconv,ren2022object}, with increased distance to the sensor, almost all approaches perform worse with sparse point clouds \cite{behley2019semantickitti}. This is especially true for classes with few examples or only a small number of points in a single scan, such as motorcyclists, trucks and poles. The task becomes very hard even for human eyes. However, in realistic situations, such as the autonomous driving scenario, accurate segmentation of sparse objects can be of great importance. For example, an object with few points, such as pedestrians and bicyclists, could be seriously affected by incorrect segmentation and can eventually lead to crashing into curbs and other road traffic accidents.



In recent years, several studies have been proposed to address the distance-dependent sparsity problem. For example, SqueezeSeg \cite{wu2019squeezesegv2} was proposed to generate a denser range image, which meant that all point cloud information was retained in the range image. However, since it adopted the commonly used 2D convolution operations for feature extraction, it could not explore the 3D geometric pattern very well. The Cylinder3D framework \cite{zhu2021cylindrical} proposed a cylindrical partition and a symmetrical 3D convolution network to better explore the 3D geometric pattern and tackle the difficulties caused by sparsity and varying density. However, as the method lacked an explicit investigation of feature representation learning for sparse objects, the segmentation performance on these objects was still less than satisfactory. PVKD \cite{hou2022point} proposed a point-to-voxel knowledge distillation approach for model compression. To enhance the distillation efficacy of the minority classes and distant objects, a difficulty-aware sampling strategy was employed to more frequently sample these hard classes. However, there were also shortcomings in this method. One of them was that there may be many background points in the supervoxels, which could cause disturbance in distillation learning for the hard classes. Another problem was that knowledge distillation was only performed in a single scan, lacking the exploration of sequential information.

Therefore, to boost the segmentation performance of instances with only a small number of points in a single scan, we propose a novel multi-to-single knowledge distillation framework for the 3D point cloud semantic segmentation task. Specifically, we try to generate more dense points for these hard instances by combining multi-scans into a single large point cloud. However, since the sequential information cannot be obtained in inference, the enhanced point cloud is only used as a supervision for better feature representation learning in model training. To this aim, we propose a simple yet effective priori-based sparse fusion strategy to make the model focus on the hard classes naturally and reduce the computational cost at the same time: instead of fusing all the points of multiple past scans directly, only the instances belonging to the previously defined classes (hard classes) are fused. We adopt point cloud registration technology to obtain more precise position alignment and accurate fusion results. For knowledge distillation, we leverage a multilevel distillation framework for knowledge translation, i.e., feature representation distillation, logit distillation, and affinity distillation. For affinity distillation, we develop a novel instance-aware affinity distillation method to capture high-level structural data more effectively for each instance, and the distillation efficiency for the hard classes can also be enhanced in the training procedure.

To evaluate the effectiveness of our proposed method, we conduct experiments on the SemanticKITTI dataset\cite{behley2019semantickitti}, and the results on both the validation and test sets demonstrate that our algorithm can outperform the baseline method by a large margin. We also conduct several ablation studies to examine the efficacy of each component. Qualitative and quantitative results are reported with a detailed description, and an analysis is also given for future works.

In summary, our contributions include the following:

1.	We propose a novel multi-to-single knowledge distillation framework for 3D point cloud semantic segmentation.

2.	We propose a simple yet effective priori-based sparse fusion strategy to make the model focus on hard classes naturally and enhance the distillation efficacy in training. 

3.	Multilevel knowledge distillation is adopted to fully exploit the information in the sequences, and a novel instance-aware affinity distillation is further proposed to better capture the high-level structural knowledge in each instance.

4.	We conduct experiments on the SemanticKITTI dataset, and superior performance is achieved over the baseline method by a significant margin.

\section{RELATED WORK}

\subsection{LiDAR Semantic Segmentation}

3D point cloud semantic segmentation is a fundamental task for the navigation of autonomous vehicles. The distance-dependent sparsity of point clouds presents a great challenge to the semantic segmentation task. SqueezeSeg \cite{wu2019squeezesegv2} proposed generating a denser range image by exploiting the way the rotating scanner captures the point cloud data, so the distance-dependent sparsity problem could be partially solved. 3DMiniNet \cite{alonso20203d} proposed learning a 2D representation from the raw points through a projection operation, which could effectively extract local and global information from the 3D data, showing promising and effective results. Cylinder3D \cite{zhu2021cylindrical} proposed a new framework to better explore the 3D geometric pattern and tackle these difficulties caused by sparsity and varying density through cylindrical partition and asymmetrical 3D convolution networks. (AF)2-S3Net \cite{cheng20212} demonstrated a multi-branch attentive feature fusion module in the encoder and an adaptive feature selection module in the decoder that could simultaneously capture and emphasize fine details for smaller instances while focusing on global contexts embodied in larger instances. In \cite{xian2022location}, based on the observation that the distribution of
objects is severely biased, it proposed a location-guided feature module to extract features with input-dependent convolutions in different regions. AMVNet \cite{liong2020amvnet} proposed a modular-and-hierarchical late fusion approach with an assertion-guided sampling strategy, where features of the uncertain points are sampled to a point head for more accurate predictions. RPVNet \cite{xu2021rpvnet} proposed a deep fusion framework with a gated fusion module that aimed to utilize different view advantages and alleviate their own shortcomings in fine-grained segmentation tasks. Although the above models have shown impressive performance on various benchmarks, the performance of classes with few points is still far from satisfactory. By combining multi-scans into a single large point cloud, the point representation of objects can be enhanced. However, how to make use of sequential information to boost the segmentation performance on these objects with few points has not been sufficiently exploited.

\subsection{Knowledge Distillation}

Knowledge distillation was introduced in the seminal work \cite{hinton2015distilling}, which aimed to minimize the KL divergence of soft class probabilities between the teacher and the student model. Knowledge distillation has been considered an effective approach for both model compression and model accuracy boosting.

SKD \cite{liu2019structured} presented two structured knowledge distillation strategies, pair-wise distillation and holistic distillation, and the pair-wise and high-order consistency was enforced between the outputs of the compact and cumbersome networks.
GID \cite{dai2021general} introduced relation-based knowledge for distillation on object detection tasks and further integrated it with response-based and feature-based knowledge, making the student model even surpass the teacher model.
PVKD \cite{hou2022point} proposed a point-to-voxel knowledge distillation approach and a difficulty-aware sampling strategy to enhance the distillation efficacy, specifically in hard cases. 
In \cite{he2019knowledge}, they proposed a new knowledge distillation method by reinterpreting the output from the teacher network to a re-represented latent domain, which made student model learning much easier, and it also proposed an affinity distillation module to help the student network capture long-term dependencies from the teacher network.
WKD \cite{zhang2022wavelet} proposed a novel knowledge distillation method that decomposed the images into different frequency bands with discrete wavelet transformation and then only distilled the high-frequency bands.
Although previous distillation approaches have shown excellent performance in broad applications in machine learning, distillation between a single scan and sequential information has not been fully explored. To the best of our knowledge, we are the first to propose the multi-to-single scan point cloud knowledge distillation.

\begin{figure*}[h]
\centering
\includegraphics[width=1\linewidth]{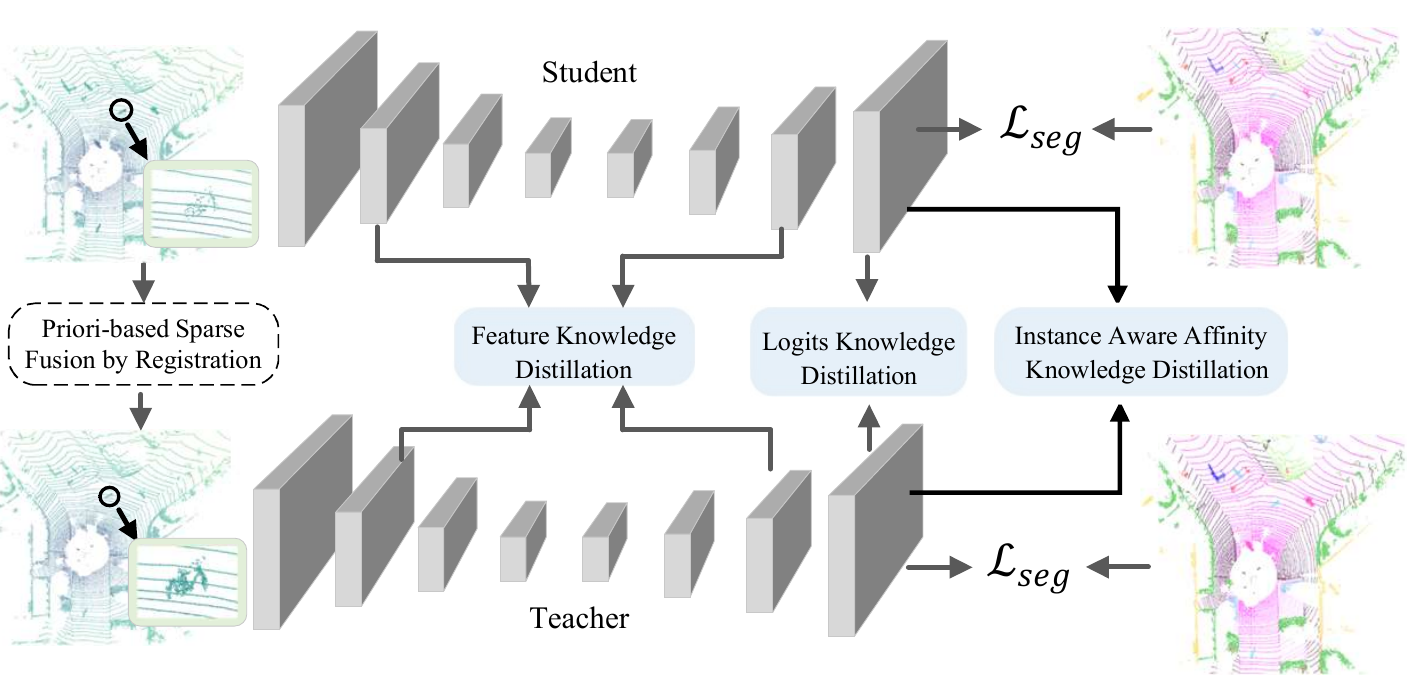}
\caption{The pipeline of our proposed multi-to-single knowledge distillation framework. There are two branches in our framework. One is the single-scan segmentation pipeline, and the other is the multi-scan fused segmentation pipeline. It should be noted that only the instances belonging to the previously defined hard classes are fused, and the points of multi-scans are precisely aligned by the point cloud registration algorithm. We leverage a multilevel knowledge distillation framework to make knowledge distillation more sufficient. Furthermore, an instance-aware affinity distillation approach is proposed to better capture high-level structural knowledge and enhance distillation efficacy for hard classes.}

\label{fig:pipeline2}
\end{figure*}

\section{PROPOSED METHOD}

In this section, we first present the details of our proposed priori-based sparse fusion strategy, and then the multilevel knowledge distillation approach is introduced. The overview of our proposed framework is shown in Figure \ref{fig:pipeline2}.

\begin{figure}[h]
\centering
\includegraphics[width=.925\linewidth]{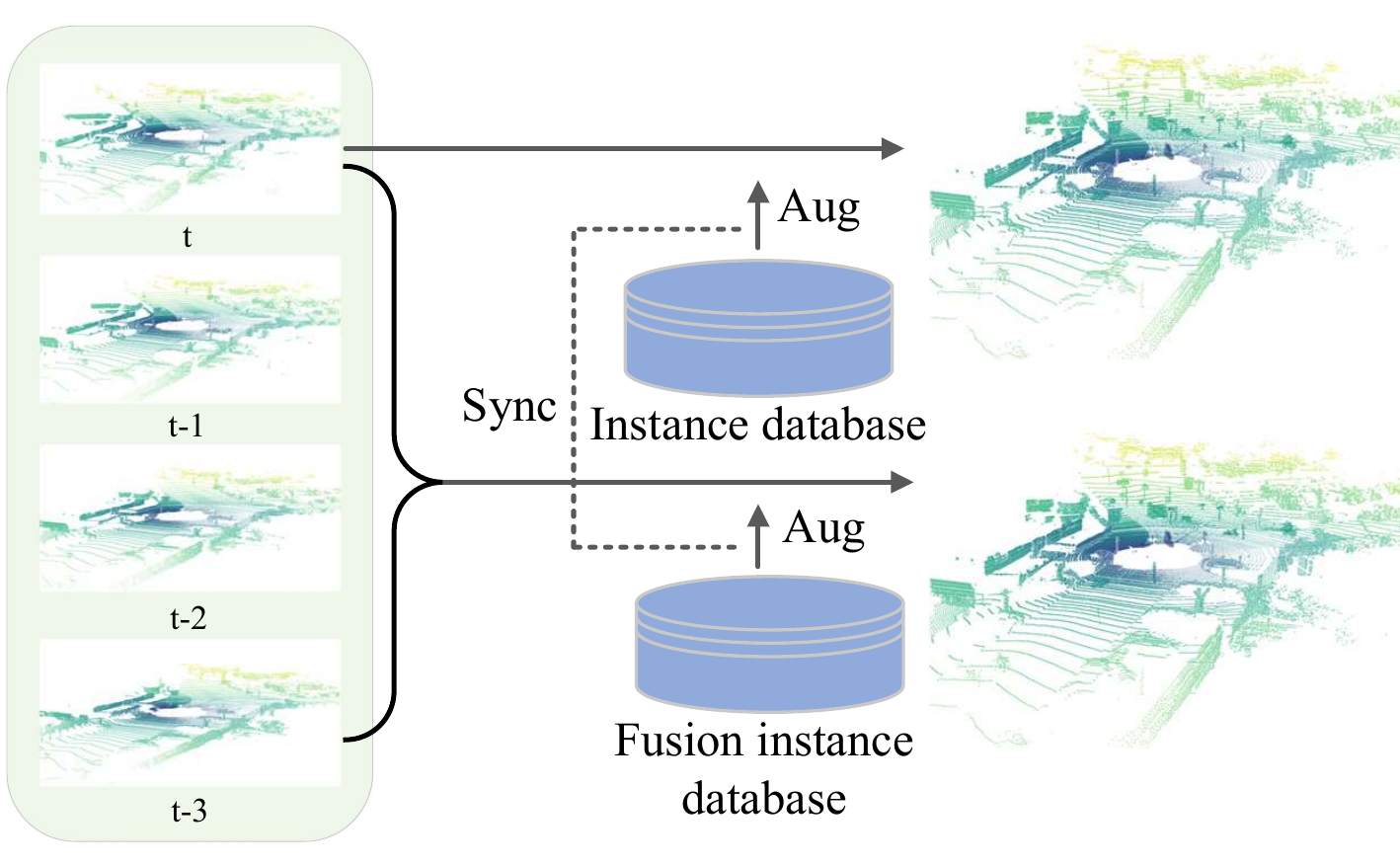}
\caption{Our copy-paste instance augmentation pipeline. For considerations of computational efficiency, first, data preprocessing is conducted for the whole dataset. Then the instances in the single scan and the corresponding fused instances are stored separately. During training, the instances in both the instance database and the fusion instance database are sampled synchronously and then pasted to the single-scan and multi-fused scans, respectively.}
\label{fig:pre_fusion}
\end{figure}

%
\subsection{Multi-scan Fusion }

\subsubsection{Priori-based Sparse Fusion} \label{subsec:priori}
By exploiting information from a sequence of multiple past scans, the segmentation performance of the current scan can be improved. However, in the standard multi-scan fusion method, the adjacent scans are simply combined by the global pose translation. Such a simple combination approach has one obvious problem, i.e., the fusion result of the current scan contains a large number of point clouds, and large quantities of computational resources are needed. In fact, in the semantic segmentation task, the performance for some ground or structure categories is already fairly good, so the sequential information is of little help to the segmentation results of these categories. 

To address the abovementioned problem, we improve the conventional multi-scan fusion into a sparse fusion based on a priori. Specifically, instead of fusing all the points of past multi-scans directly, only the points belonging to the previously defined classes are fused. The classes with with a lower IoU are selected as the candidate classes to be enhanced by exploiting the sequential information. Our approach has two obvious advantages compared to conventional multi-scan fusion methods. One is the reduced computational cost of the fused point cloud, and the other is to force the model to focus on difficult samples in the knowledge distillation process.

For objects without instance IDs in the SemanticKITTI dataset \cite{aygun20214d}, such as traffic signs, we proposed a simple method, for instance, ID generation. In this method, we only try to generate an instance ID for each object using semantic labels. However, as there may be many objects belonging to the same class in each scan, the algorithm must be carefully designed. Next, we provide a detailed description of the instance ID generation method. Specifically, we first filter the point cloud based on semantic labels, and only points that belong to the specific class are preserved. Then, the farthest point sampling is performed on the preserved points. If the distance between the points sampled currently and previously is less than a certain threshold, the sampling operation is stopped. Next, we take the sampled points as key points, and then a distance-based clustering algorithm is performed with these sampled points. Finally, the unique instance IDs were assigned to each cluster. The instance ID generation pipeline is shown in Figure \ref{fig:ins_gen}.

To further boost the performance of the hard class, we also adopt the copy-paste instance augmentation approach \cite{ghiasi2021simple,Zhou2021PanopticPolarNet} in our knowledge distillation framework. The augmentation pipeline is shown in Figure \ref{fig:pre_fusion}.

Another challenging problem is to tackle the moving and non-moving classes in each scan, i.e., cars and buildings. The straightforward fusion method based on the global pose translation cannot handle moving objects. To solve this problem, we separate the fusion process into two different parts: moving objects and non-moving objects, respectively. Non-moving object fusion can be simply achieved by the global pose translation and multi-scan point cloud combination. Because there is no pose translation information for each object, moving object fusion becomes more difficult. To cope with this problem, we adopt point cloud registration technologies to obtain more precise fusion results, and the details of the fusion by registration approach are presented in the following section.

\begin{figure*}[h]
\centering
\includegraphics[width=0.95\linewidth]{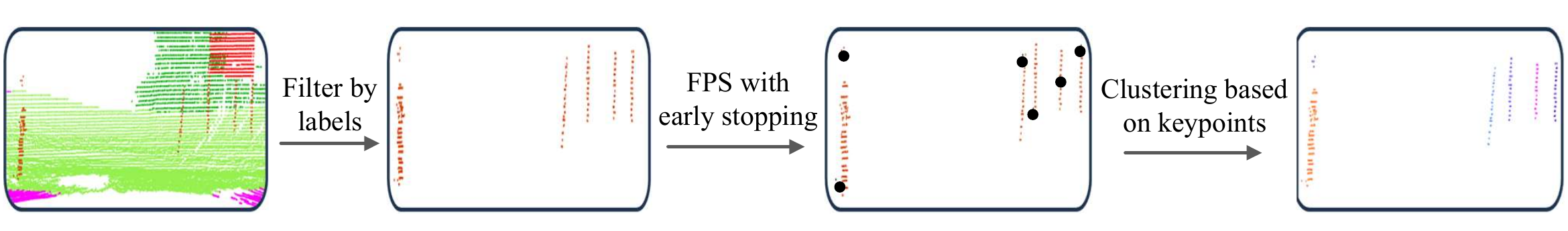}
\caption{The instance ID generation pipeline for specific classes. }
\label{fig:ins_gen}
\end{figure*}

\subsubsection{Fusion by Registration}
In the priori-based sparse fusion, one of the most straightforward ways is that for each moving instance, we simply move the point cloud of the corresponding instance in the adjacent scans to the current position. However, in this case, determining the precise location of the instance becomes crucial. We found experimentally that the commonly used centroid-based methods work well in most cases, and the centroid-based approach is also very efficient in practice. However, in regard to the sparser instances, such as the bicyclist and motorcyclist in the distance, or the car in a turn, the simple centroid-based alignment method will become fail.

To achieve a more precise fusion result, we adopt point registration technology to solve the misalignment problem in the fusion operation. Specifically, for each candidate instance in the current scan, we first gather all the corresponding instances in the adjacent scans, and then we use the centroid-based method as a fast way to provide a feasible initialization for the following registration algorithm. Finally, a more precise registration-based point cloud fusion operation is performed.

Noted that in the SemanticKITTI dataset, the instance ID is consistent in the adjacent scans, so we can use the ground truth data in the above registration alignment operation to find the same instance in the adjacent scans. In other datasets where a consistent instance ID cannot be obtained from the annotation data, one feasible way is to generate them using object-tracking methods.

\subsection{Multilevel Knowledge Distillation}
Different from the standard knowledge distillation task, which aims to facilitate the training of a lightweight student model under the supervision of a sophisticated teacher model, we try to distill the enhanced feature representation from the multi-scan combination input to the single scan input.

\subsubsection{Feature Representation Distillation}

To make the knowledge distillation from multi-scans more sufficient, we leverage a multilevel distillation framework for knowledge translation, i.e., feature representation distillation, logit distillation, and affinity distillation. For the feature-level knowledge distillation, we select the feature from two different middle layers of the segmentation network for knowledge distillation learning, i.e., the feature after the second down-sampling block and the feature after the third up-sampling block. As mentioned in subsection \ref{subsec:priori}, we aim to encourage the network to focus on the features of the predefined hard classes. Since only the classes whose segmentation performance is below a predefined threshold are fused by the priori-based sparse fusion operation, we simply constrain the knowledge distillation regions at the points of fusion areas. Finally, referring to the methods described in \cite{du2021ago}, we adopt the smooth-L1 distance between the student and teacher model outputs, which is directly optimized for feature-level knowledge distillation.
\begin{equation}
\scriptsize
\begin{aligned}
\label{con:FD}
& \mathcal{L}_{FD} = \\ &\begin{cases} \frac{1}{Nf_c} \sum\limits_{i=1}^{N} \sum\limits_{k=1}^{f_c} \frac{1}{2T} \|F_{(i,k)}\!-\!f_{(i,k)}\|^2 & \text {} \|F_{(i,k)}\!-\!f_{(i,k)}\|<T \\ \frac{1}{Nf_c} \sum\limits_{i=1}^{N} \sum\limits_{k=1}^{f_c} \frac{1}{2T} \|F_{(i,k)}\!-\!f_{(i,k)}\|\!-\!0.5T^2 & \text { otherwise }\end{cases},
\end{aligned}
\end{equation}
where $N$ is the number of sampled hard points, $f_c$ is the channels of the feature, $F_{(i,k)}$ and the $f_{(i,k)}$ are the features of the  teacher model and the student model respectively, and $T$ is a predefined threshold parameter.

\subsubsection{Logits Distillation}
In the logits output layer, knowledge distillation is also conducted. Similar to feature representation distillation, only the specific regions are considered in the logit distillation. Following much prior work, we use the Kullback-Leibler divergence loss to minimize the KL divergence of class probabilities between the multi-scan and single-scan outputs. Soft labels are always considered to contain much more information than one-hot labels, which can play the roles of supervisory signals and regularizations at the same time. Therefore, in our experiment, we adopt the softened version of the teacher and student model logits by multiplying the logits with a temperature parameter $P$.

\begin{equation}
\begin{aligned}
\label{con:SLDl}
\mathcal{L}_{SLD} = \frac{1}{NC} \sum_{i=1}^{N} \sum_{c=1}^{C} \varphi(F_l(i,c)) \frac{\varphi(F_l(i,c))}{\varphi(f_l(i,c))},
\end{aligned}
\end{equation}
where $C$ is the channels of the logits, $\varphi()$ is the softmax function.

\begin{figure}[h]
\centering

\subfloat[PVKD]{\includegraphics[width = 0.45\linewidth]{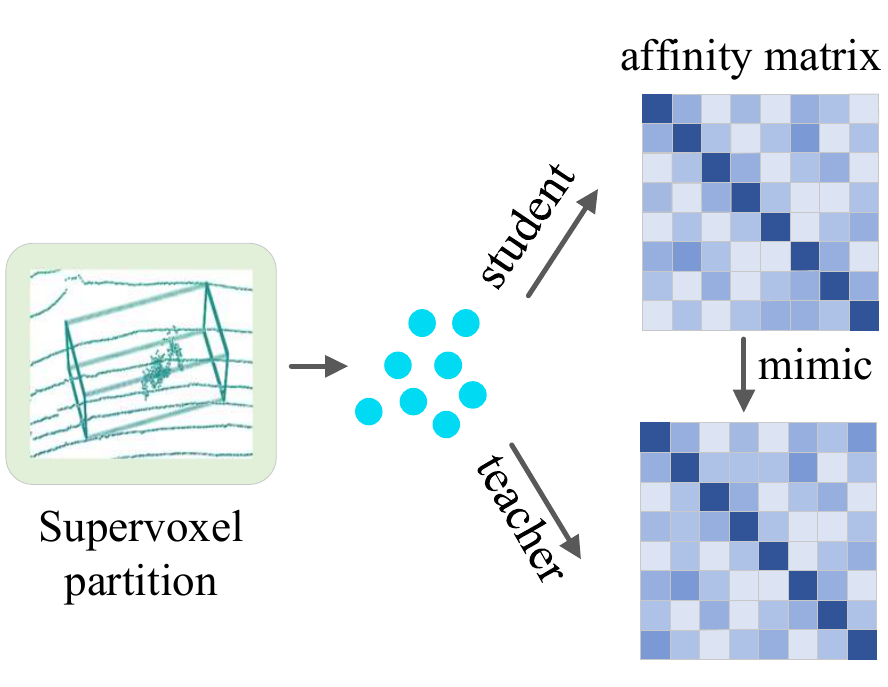}}
	\hfill
\subfloat[Ours]{\includegraphics[width = 0.47\linewidth]{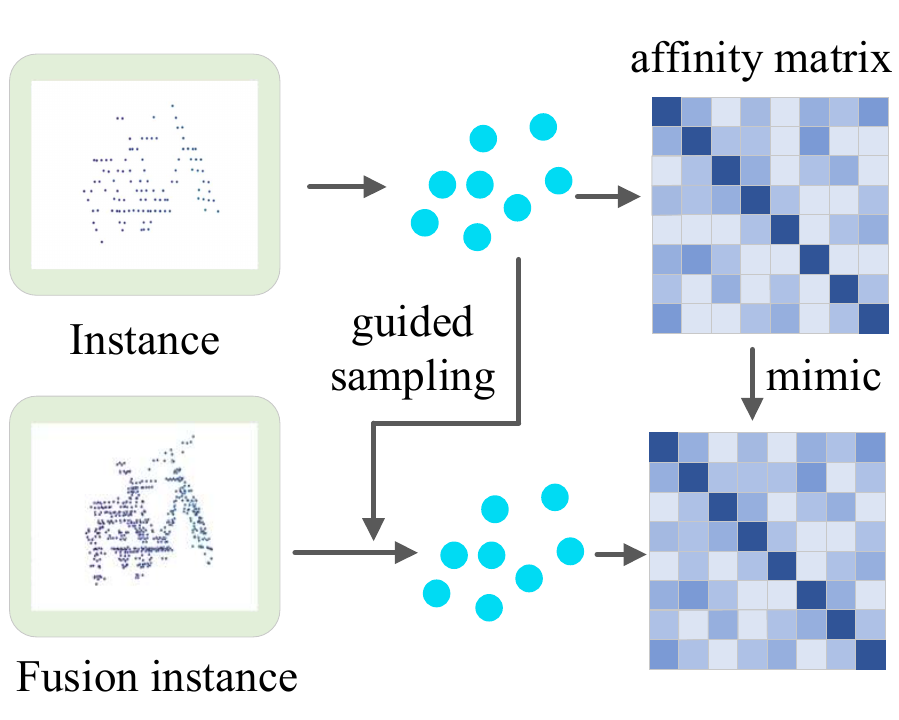}}

\caption{Comparison between our proposed instance-aware affinity knowledge distillation approach and the affinity knowledge distillation method proposed in PVKD\cite{hou2022point}. Note that the interference of background points is totally removed in our approach, and the high-level structural knowledge of each instance is captured more precisely and effectively. }
\label{fig:IAAD}
\end{figure}

\subsubsection{Affinity Distillation}
Distilling the knowledge of the output features is insufficient since it considers only individual elements while the structural information of the surrounding environment is ignored. However, structural knowledge is crucial to the 3D point cloud semantic segmentation model as the input points are unordered, especially regarding objects with only a small number of points. One possible solution is the relational knowledge distillation technology proposed in \cite{park2019relational}. Unlike pixel-wise feature representation knowledge distillation, relational knowledge distillation attempts to distill the pair-wise relationship of all point features. However, as mentioned in \cite{hou2022point}, the affinity distillation process became computationally expensive and extremely difficult to learn since the similarity matrix had too many elements. In \cite{hou2022point}, they improved the previous relational knowledge distillation approach by dividing the whole point cloud into several super voxels and only sampling part of them for the relation distillation learning. However, direct applying this method is also problematic, as we mainly focus on obtaining better feature representations of classes with only a small number of points, which is most likely to be objects such as bicyclists or motorcyclists. The supervoxel division mechanism is not quite suitable for these small objects. Hence, we develop a simple yet effective instance-aware affinity distillation strategy to overcome the above issues. We only calculate the pair-wise similarity of point clouds belonging to the same instances. As these objects always have only a small number of points, the simpler similarity matrix is computationally efficient and much easier to learn. In addition, the model can focus on each instance in the distillation process automatically. The comparison between our instance-aware affinity distillation approach and the affinity distillation proposed in PVKD is shown in Figure \ref{fig:IAAD}.

The instance-aware affinity matrix is calculated according to the following equation:

\begin{equation}
\begin{aligned}
\label{con:affmat}
A_{matrix}(i,j) = \frac{Fea(i)^TFea(j)}{\|Fea(i)\|^2\|Fea(j)\|^2} \quad i,j \in \mathcal{S}_k,
\end{aligned}
\end{equation}
where $Fea$ denotes the features of the teacher model or the student model and $\mathcal{S}_k$ is the point set of the k$th$ instance.

Then, the affinity distillation loss is formulated as:

\begin{equation}
\begin{aligned}
\label{con:IAAD}
\mathcal{L}_{IAAD} = \sum_{k=1}^{|\mathcal{S}|} \frac{1}{{|\mathcal{S}_k|}^2} \sum_{i=1}^{|\mathcal{S}_k|} \sum_{j=1}^{|\mathcal{S}_k|} \| A_m(i,j) - A_s(i,j) \|^2.
\end{aligned}
\end{equation}

\subsection{The Final Objective Function}

Our final loss function is composed of four terms, i.e., the main segmentation task loss for the point-wise prediction of both the student and teacher models, the smooth-L1 loss for the feature representation distillation, the Kullback-Leibler divergence loss for soft logits distillation, and the 12 loss for the instance-aware affinity distillation:

\begin{equation}
\begin{aligned}
\label{con:final}
\mathcal{L}_{final} = & \mathcal{L}_{seg}^{S} + \beta_{1} \mathcal{L}_{seg}^{T} \\ &+ \beta_{2}\mathcal{L}_{FD} + \beta_{3}\mathcal{L}_{SLD} + \beta_{4}\mathcal{L}_{IAAD},
\end{aligned}
\end{equation}
where $\beta_{1}$, $\beta_{2}$, $\beta_{3}$, and $\beta_{4}$ are the loss coefficients used to balance the contribution of the knowledge distillation loss to the main segmentation task loss. In our experiments, if not otherwise stated, $\beta_{1}$, $\beta_{2}$ and $\beta_{3}$ are set as 0.5, 0.01, 0.1, and 0.1.

\section{EXPERIMENTAL RESULTS}

\subsection{Datasets}
SemanticKITTI\cite{behley2019semantickitti} is a large-scale dataset for semantic scene understanding using LiDAR sequences, which is based on the KITTI Vision Benchmark and has a dense semantic annotation for the entire KITTI Odometry Benchmark, making it a standard dataset for evaluating LiDAR semantic segmentation methods. This dataset presents challenges on rare classes such as motorcyclists and other-ground due to the limited training examples. The most frequent class, ``vegetation", has $4.82 \times 10^7$ times more points than the least frequent class, ``motorcyclist", which is heavily imbalanced.  In our experiments, we defined the bicycle, motorcycle, truck, other-vehicle, person, bicyclist, motorcyclist, and traffic-sign as the hard classes, and we adopt the method in \cite{yang2019quaternion} for point cloud registration.



\subsection{Main Results}

As official guidance suggests, we use mean intersection-over-union (mIoU) over all classes as the evaluation metric. The metric can be formalized as follows:

\begin{equation}
\begin{aligned}
\label{con:mIoU}
m I o U=\frac{1}{n} \sum_{c=1}^n \frac{T P_c}{T P_c+F P_c+F N_c},
\end{aligned}
\end{equation}
where $TP_{c}$ denotes the number of true positive points for class $c$,
$FP_c$ denotes the number of false positives, and $FN_c$ is the number
of false negatives. As the name suggests, the final IoUs are calculated for each class separately and then the mean is taken.

\begin{table*}[ht]
  \centering
  \scriptsize
  \caption{Quantitative comparison of Cylinder3D trained with our framework. The results are reported in terms of the mIoU on the SemanticKITTI test set. $*$ represents the result reproduced by the officially released code and models. From top to
bottom, the methods are grouped into point-based, projection-based, and multi-view fusion models.}
    \setlength{\tabcolsep}{1.2mm}{
    \begin{tabular}{l|lllllllllllllllllll|l}
    \hline
    & & & & & & & & & & & & & & & & & & & &\\
    & & & & & & & & & & & & & & & & & & & &\\
    \multicolumn{1}{l|}{Methods}  & {\rotatebox{70}{car}} & {\rotatebox{70}{bicycle\hspace{-1cm}}} & {\rotatebox{70}{motorcycle\hspace{-1cm}}} & {\rotatebox{70}{truck\hspace{-1cm}}} & {\rotatebox{70}{other-vehicle\hspace{-1cm}}} & {\rotatebox{70}{person\hspace{-1cm}}} & {\rotatebox{70}{bicyclist\hspace{-1cm}}} & {\rotatebox{70}{motorcyclist\hspace{-1cm}}} & {\rotatebox{70}{road}} & \multicolumn{1}{l}{\rotatebox{70}{parking}} & {\rotatebox{70}{sidewalk}} & {\rotatebox{70}{other-ground\hspace{-1cm}}} & {\rotatebox{70}{building\hspace{-1cm}}} & {\rotatebox{70}{fence\hspace{-1cm}}} & {\rotatebox{70}{vegetation\hspace{-1cm}}} & {\rotatebox{70}{trunk\hspace{-1cm}}} & {\rotatebox{70}{terrain\hspace{-1cm}}} & {\rotatebox{70}{pole\hspace{-1cm}}} & {\rotatebox{70}{traffic-sign\hspace{-1cm}}} & \multicolumn{1}{l}  {mIoU} \\    \hline
    \multicolumn{1}{l|}{LatticeNet \cite{rosu2019latticenet}} &   88.6 & 12.0 & 20.8 & 43.3 & 24.8 & 34.2 & 39.9 & 60.9 & 88.8 & 64.6 & 73.8 & 25.5 & 86.9 & 55.2 & 76.4 & 67.9 & 54.7 & 41.5 & 42.7 & 51.3 \\
    \multicolumn{1}{l|}{PointNL \cite{cheng2020cascaded}} &    92.1 & 42.6 & 37.4 & 9.8 & 20.0 & 49.2 & 57.8 & 28.3 & 90.5 & 48.3 & 72.5 & 19.0 & 81.6 & 50.2 & 78.5 & 54.5 & 62.7 & 41.7 & 55.8 & 52.2\\

    \multicolumn{1}{l|}{RandLa-Net \cite{hu2020randla}} &  94.2 & 26.0 & 25.8 & 40.1 & 38.9 & 49.2 & 48.2 & 7.2 & 90.7 & 60.3 & 73.7 & 20.4 & 86.9 & 56.3 & 81.4 & 61.3 & 66.8 & 49.2 & 47.7 & 53.9 \\
    
    \multicolumn{1}{l|}{KPConv \cite{thomas2019kpconv}} &    96.0 & 30.2 & 42.5 & 33.4 & 44.3 & 61.5 & 61.6 & 11.8 & 88.8 & 61.3 & 72.7 & 31.6 & 90.5 & 64.2 & 84.8 & 69.2 & 69.1 & 56.4 & 47.4 & 58.8\\
    \hline
    \multicolumn{1}{l|}{SqueezeSegV2 \cite{wu2019squeezesegv2}} & 82.7  & 21.0    & 22.6  & 14.5  & 15.9  & 20.2  & 24.3  & 2.9   & 88.5  & 42.4  & 65.5  & 18.7  & 73.8  & 41.0    & 68.5  & 36.9  & 58.9  & 12.9  & 41.0    & 39.6 \\
    
    \multicolumn{1}{l|}{RangeNet53 \cite{milioto2019rangenet++}} 
     & 91.4  & 25.7  & 34.4  & 25.7  & 23.0    & 38.3  & 38.8  & 4.8   & 91.8  & 65.0    & 75.2  & 27.8  & 87.4  & 58.6  & 80.5  & 55.1  & 64.6  & 47.9  & 55.9  & 52.2 \\
    
    \multicolumn{1}{l|}{3D-MiniNet-KNN \cite{alonso20203d}} &  90.5 & 42.3 & 42.1 & 28.5 & 29.4 & 47.8 & 44.1 & 14.5 & 91.6 & 64.2 & 74.5 & 25.4 & 89.4 & 60.8 & 82.8 & 60.8 & 66.7 & 48.0 & 56.6   & 55.8 \\
    
    \multicolumn{1}{l|}{SqueezeSegV3 \cite{xu2020squeezesegv3}} &  92.5 & 38.7 & 36.5 & 29.6 & 33.0 & 45.6 & 46.2 & 20.1 & 91.7 & 63.4 & 74.8 & 26.4 & 89.0 & 59.4 & 82.0 & 58.7 & 65.4 & 49.6 & 58.9 & 55.9 \\
    \multicolumn{1}{l|}{SalsaNext \cite{cortinhal2020salsanext}} &     91.9 & 48.3 & 38.6 & 38.9 & 31.9 & 60.2 & 59.0 & 19.4 & 91.7 & 63.7 & 75.8 & 29.1 & 90.2 & 64.2 & 81.8 & 63.6 & 66.5 & 54.3 & 62.1 & 59.5\\
    
     \multicolumn{1}{l|}{KPRNet \cite{kochanov2020kprnet}} &  95.5  & 54.1  & 47.9  & 23.6  & 42.6  & 65.9  & 65.0  & 16.5  & 93.2  & 73.9  & 80.6  & 30.2  & 91.7  & 68.4  & 85.7  & 69.8  & 71.2  & 58.7  & 64.1 & 63.1\\    
    
    \hline
    
    \multicolumn{1}{l|}{MVLidarNet \cite{chen2020mvlidarnet}} & 87.1 & 34.9 & 32.9 & 23.7 & 24.9 & 44.5 & 44.3 & 23.1 & 90.3 & 56.7 & 73.0 & 19.1 & 85.6 & 53.0 & 80.9 & 59.4 & 63.9 & 49.9 & 51.1 & 52.5 \\
    
     \multicolumn{1}{l|}{MPF \cite{alnaggar2021multi}} &  93.4 & 30.2 & 38.3 & 26.1 & 28.5 & 48.1 & 46.1 & 18.1 & 90.6 & 62.3 & 74.5 & 30.6 & 88.5 & 59.7 & 83.5 & 59.7 & 69.2 & 49.7 & 58.1 & 55.5 \\
    
    \multicolumn{1}{l|}{TORNADONet \cite{gerdzhev2021tornado}} &  94.2  & 55.7  & 48.1  & 40.0  & 38.2  & 63.6  & 60.1  &  34.9  & 89.7  & 66.3  & 74.5  & 28.7  & 91.3  & 65.6  & 85.6  & 67.0  & 71.5  & 58.0 &  65.9 & 63.1 \\

    \multicolumn{1}{l|}{AMVNet \cite{liong2020amvnet}} &  96.2 & 59.9 & 54.2 & 48.8 & 45.7 & 71.0 & 65.7 & 11.0 & 90.1 & 71.0 & 75.8 & 32.4 & 92.4 & 69.1 & 85.6 & 71.7 & 69.6 & 62.7 & 67.2 & 65.3 \\
    
    \multicolumn{1}{l|}{GFNet \cite{qiu2022gfnet}} & 96.0 & 53.2 & 48.3 & 31.7 & 47.3 & 62.8 & 57.3 & 44.7 & 93.6 & 72.5 & 80.8 & 31.2 & 94.0 & 73.9 & 85.2 & 71.1 & 69.3 & 61.8 & 68.0  & 65.4 \\
    
    \hline

    \multicolumn{1}{l|}{PolarNet \cite{zhang2020polarnet}} &  93.8 & 40.3 & 30.1 & 22.9 & 28.5 & 43.2 & 40.2 & 5.6 & 90.8 & 61.7 & 74.4 & 21.7 & 90.0 & 61.3 & 84.0 & 65.5 & 67.8 & 51.8 & 57.5  & 54.3 \\

    \multicolumn{1}{l|}{PolarNet+Ours} &  93.5 & 45.9 & 36.3 & 27.6 & 34.9 & 55.0 & 51.4 & 15.8 & 91.1 & 64.7 & 73.8 & 26.1 & 92.5 & 67.0 & 84.6 & 63.4 & 67.4 & 50.7 & 59.5 & 58.0 \\
    
    \multicolumn{1}{l|}{Cylinder3D* \cite{zhu2021cylindrical}} &  96.7 & 60.1 & 57.4 & 43.2 & 49.6 & 70.0 &  65.1 &  12.0 & 91.6 & 64.6 & 76.0 & 24.3 & 90.0 & 63.4 & 84.8 & 70.7 & 67.6 & 62.0 & 64.0 & 63.9\\    

    \multicolumn{1}{l|}{Cylinder3D+Ours} & 96.4 & 60.8 & 54.8 & 42.8 & 51.2 & 69.1&  67.8&  34.8 & 92.2 & 66.5 & 76.7 & 30.4 & 91.1 & 65.7 & 85.5 & 69.8 & 68.6 & 60.7 & 61.0 & 65.6\\    
    \hline
    \end{tabular}}
  \label{tab:exp}
\end{table*}

To verify the effectiveness of our method, we apply our method to both the PolarNet \cite{zhang2020polarnet} and Cylinder3D \cite{zhu2021cylindrical} baselines. The experimental results are shown in Table \ref{tab:exp}. Note that the Cylinder3D* represents the result reproduced by the model released by the author without test augmentation. For a fair comparison, our model is only trained on the training set. It can be seen that we surpass the PolarNet baseline by 3.7\% mIoU, and surpass the Cylinder3D baseline by 1.7\% mIoU. Specifically, for hard classes such as bicyclist and motorcyclist, the performance gain of our proposed multi-to-single knowledge distillation framework becomes more significant, we surpass the PolarNet baseline by 11.2\% and 10.2\%, and surpass the Cylinder3D baseline by 2.7\% and 22.8\% respectively.

\subsection{Ablation Study}

\begin{table}[t]   
    \centering
    \caption{Ablation study of each component on the final performance on the SemanticKITTI validation set.}  
\begin{tabular}{c|c|c|c}
\hline \multirow{1}{*}{\text {Cylinder3D}} &  \multirow{1}{*}{\text {PWKD}} & \multirow{1}{*}{\text {IAAD}}  &  \multirow{1}{*}{\text {$\mathrm{mIoU}$}} \\
\hline $\sqrt{ }$ & & &   $64.1$ \\
$\sqrt{ }$& $\sqrt{ }$ &  &   $65.6$ \\
$\sqrt{ }$& $\sqrt{ }$ & $\sqrt{ }$  & $66.2$ \\
\hline
\end{tabular}
    \label{tab:Ablation}
\end{table}

We performed ablation studies to examine the efficacy of the major components of the proposed method. The ablation results are shown in Table \ref{tab:Ablation}. We reproduce the results of Cylinder3D by the officially released code. PWKD denotes the point-wise knowledge distillation, including the feature distillation and the logit distillation, IAAD denotes the instance-aware affinity distillation.



\subsection{Visualization}

Finally, we also visualize the performance gain of our approach in different scans and mainly focus on two of the hard classes, including truck and bicyclist. The result is shown in Figure \ref{fig:vis}. Note that in scene A, our model gives better results on the prediction of the truck segmentation. In scene B, the baseline model misclassified a bicyclist as a person, as these two classes are very similar; however, our method gives the correct predictions.

\begin{figure}[h]
\centering

\subfloat[Scene A]{\includegraphics[width = .9\linewidth]{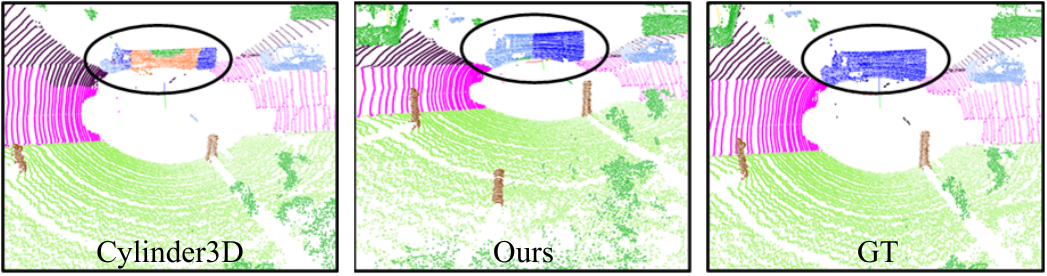}}
\vspace{0.1cm}
\subfloat[Scene B]{\includegraphics[width = .9\linewidth]{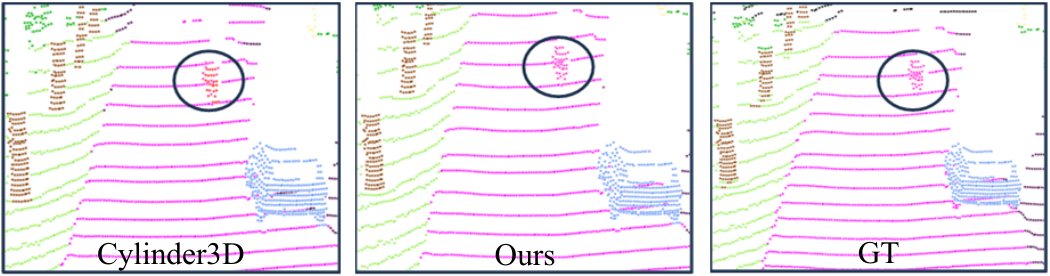}}
	
\caption{Visualization comparison of the baseline (Cylinder3D) and our proposed method on the SemanticKITTI dataset.}
\label{fig:vis}
\end{figure}

\section{CONCLUSIONS}

In this work, we have developed a simple but effective multi-to-single knowledge distillation framework for the 3D point cloud semantic segmentation task. By training the model with supervision from multiple adjacent scans combined, the segmentation performance of the single scan can be improved. We adopt the priori-based sparse fusion strategy to fuse only the predefined hard classes, which can make the model focus on the hard classes naturally and enhance the distillation efficacy in the training procedure. We also propose an instance-aware affinity distillation to better capture the high-level structural knowledge in each object in knowledge distillation. The multi-to-single knowledge distillation framework can also be considered as a plug-in component and can be integrated into the recently developed point cloud semantic segmentation approaches. We conduct experiments on the SemanticKITTI dataset and the results demonstrate that our algorithm can outperform the baseline method by a significant margin. One of our future works is to extend our proposed approach to other 3D point cloud understanding tasks, such as 3D object detection.










\clearpage
\bibliographystyle{IEEEtran}
\bibliography{sample}

\end{document}